\documentclass[letterpaper, 10 pt, conference]{ieeeconf}  

\IEEEoverridecommandlockouts                              

\usepackage{graphics} 
\usepackage{epsfig} 
\usepackage{mathptmx} 
\usepackage{times} 
\usepackage{amsmath} 
\usepackage{amssymb}  
\usepackage[dvipsnames]{xcolor}
\usepackage{cite}
\usepackage{subcaption}

\newcommand{\OURS}{SceneGrasp}
\newcommand{\SSGNET}{SceneGrasp-Net}
\newcommand{\SSGAE}{ScaleShapeGrasp-AE}
\newcommand{\FPS}{30}

\newcommand{\secref}[1]{Section~\ref{#1}}
\newcommand{\tabref}[1]{Table~\ref{#1}}

\newcommand{\figref}[1]{Fig.~\ref{#1}}

\title{\LARGE \bf
Real-time Simultaneous Multi-Object 3D Shape Reconstruction, 6DoF Pose Estimation and Dense Grasp Prediction
}

\author{Shubham Agrawal, Nikhil Chavan-Dafle, Isaac Kasahara, Selim Engin, Jinwook Huh, Volkan Isler
\thanks{All authors are with the Samsung AI Center NY, 837 Washington St, New York, NY 10014.}%
}

\begin{document}

\maketitle
\thispagestyle{empty}
\pagestyle{empty}

\begin{abstract}
Robotic manipulation systems operating in complex environments rely on perception systems which provide information about the geometry (pose and 3D shape) of the objects in the scene along with other semantic information such as object labels. This information is then used for choosing the feasible grasps on relevant objects.  
In this paper, we present a novel method to provide 
this geometric and semantic information of all objects in the scene as well as feasible grasps on those objects simultaneously. The main advantage of our method is its speed as it avoids sequential perception and grasp planning steps. 
%
With detailed quantitative analysis we show that our method delivers competitive performance compared to the state-of-the-art dedicated methods for object shape, pose, and grasp predictions, while providing fast inference at \FPS{} frames per second speed.
\end{abstract}


\section{INTRODUCTION}



During task-driven robotic manipulation, the robot needs to extract various types of information about objects in the scene.
A high-level planner needs \emph{semantic information} in the form of object labels for example to choose which object to manipulate, or whether a side or a top-down grasp is necessary. \emph{Geometric information} regarding the shape or pose of the objects is used for grasp planning.
For example, a robot loading a dishwasher needs to place cups-and-bowls, dishes, and utensils in specific sections and in specific orientations in the dishwasher racks. To selectively grasp a bowl and load it appropriately in the dishwasher, the robot needs to identify the bowl in the scene and localize it. The understanding of the full 3D geometry of the object dictates where the robot can grasp the object and moreover how to place the object in the task.

\begin{figure}
    \centering
    \includegraphics[width=1.0\columnwidth]{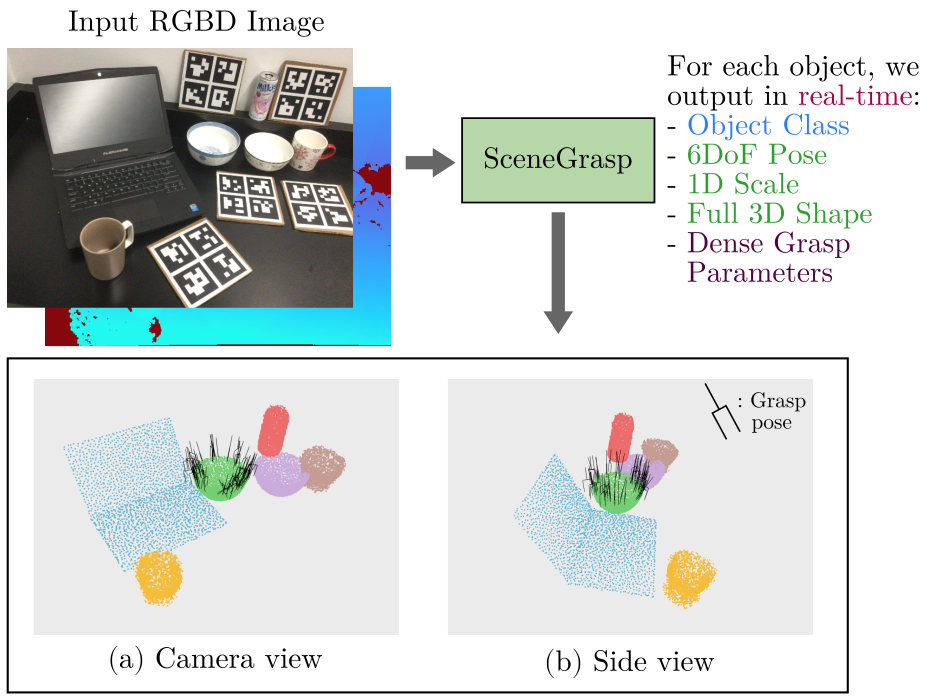}
    \caption{Given an RGBD image as input, \OURS{} detects all novel object instances, their object class, 6DoF pose, 1D scale, full 3D shape, and dense grasp parameters in real-time. (a) shows the predictions in camera frame. (b) shows the side view predictions where we see full 3D shape reconstruction and predicted grasps on unobserved regions. For clarity, we show dense grasp predictions only for one object.}
    \label{fig:teaser}
\end{figure}
    
Semantic scene understanding, including the detailed information of the object categories, object poses, and object geometries, plays an important role in guiding the robot actions such as grasp and motion planning.  However, scene understanding and action planning are often studied separately and the interdependence of the two is only recently studied~\cite{jiang2021synergies, zeng2020transporter, chavan2021object}.
In this paper, we present \textbf{\OURS{}}, a novel method to simultaneously infer geometric shape-grasp predictions and semantic information regarding all objects in the scene.
Specifically, given an RGBD image of a cluttered scene, \OURS{} outputs semantic segmentation information at the scene level, along with the pose, full 3D geometry and feasible grasps to manipulate each object in the scene.


\OURS{} has a number of desirable properties:
Generating 3D reconstructions in the object frame yields very accurate 3D models. Estimating the pose of the object allows us to put the 3D models in the camera frame which allows for action planning directly in the camera frame. In this regard, \OURS{} builds on  CenterSnap~\cite{irshad2022centersnap} and adds the capability of outputting dense grasp feasibility maps for each object. This is non-trivial because grasp feasibility depends on the scale of the object with respect to the gripper which \OURS{} learns explicitly during training. Finally, \OURS{} provides scene understanding and dense grasp estimation at \emph{\FPS{} frames per second} (FPS) which is much faster than traditional pipeline of predicting object segmentation, pose, shape, and grasp prediction sequentially.


\begin{figure*}[ht!]
    \centering
    \includegraphics[width=\linewidth]{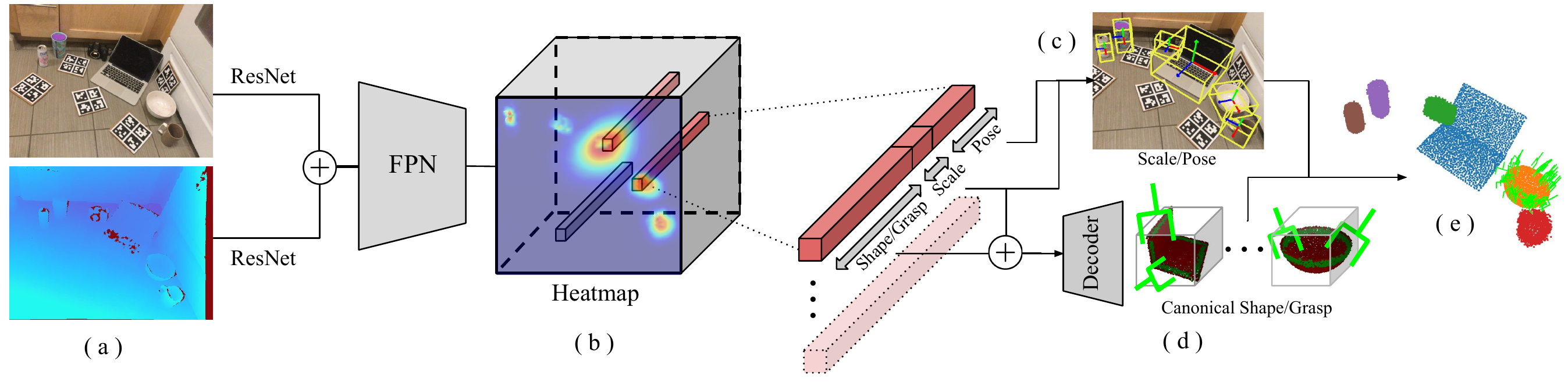}
    \caption{\textbf{\SSGNET{} Overview:} \SSGNET{} at inference time takes in an input RGB image and the corresponding depth image (a), and passes them through a ResNet architecture before concatenating the features together. These features are then passed through a Feature Pyramid Network (FPN) to obtain the heatmap (b). Embeddings are extracted at the peaks of the heatmap, and each embedding is then separated into 3 parts.  The shape/grasp embedding is concatenated with the scale embedding and passed to the our decoder to predict unit canonical shape and dense grasps (d).  This point cloud containing shape/grasp is then scaled and transformed to the camera frame using the pose/scale embedding (c) to obtain our final prediction (e).}
    \label{fig:method}
\end{figure*}

\OURS{} is trained in two steps: 
First, we train an object-scale dependent auto-encoder for object-frame shape reconstruction and grasp prediction. Second, we train a network to predict the semantic segmentation, object poses, object scales, and encodings for shape-grasp decoder. These two networks work together to generate object meta information, shapes, and grasp predictions in unit canonical object frame. Using the estimated object scales and poses we transform the object shapes and grasps to the camera/robot frame as shown in \figref{fig:teaser}. 


With rigorous quantitative evaluation we demonstrate that \OURS{} is able to simultaneously estimate object meta information and shape-and-grasp predictions accurately and in real-time (\FPS{} FPS). 

To summarize, our key contributions are: 
\begin{itemize}
    \item We present object-level scene understanding (object classification, reconstruction, and pose estimation), and dense grasp estimation method for multiple objects from a single view RGBD image in a single feed-forward pass manner.
    \item We quantitatively show competitive scene understanding and grasp success performance with traditional multi-stage pipelines while being much faster.
\end{itemize}

We believe that real-time object-level scene understanding and grasp prediction capabilities provided by our work will pave the way for reactive task-aware object manipulation in cluttered environments.
\section{Related Work}

\subsection{3D Shape reconstruction and Pose Estimation}
Scene understanding provides foundation for many down-the-stream tasks. Advances in machine learning has contributed to significant improvements to image-level scene understanding tasks such as image segmentation, object detection, and object tracking~\cite{faster_rcnn, mask_rcnn, wang2019realtime}. The high accuracy and more importantly the fast computation speed of these methods have led to wide-spread use of these technologies in everyday products such as cellphones, cameras, and also for robots. 

The 3D scene understanding involving object geometry and pose however, remains a challenging and active topic for research. While many of the pose prediction methods limit to objects of known 3D geometry~\cite{xiang2018posecnn, wang2019densefusion, tremblay2018corl:dope}, recent object reconstruction methods extend the application to novel objects within known categories or in some cases to novel object categories as well\cite{wang18p2m, mescheder19occnet, mitchel20hof, park19deepsdf}. Extensive research on object shape reconstruction has demonstrated that object-frame reconstruction methods provide accurate reconstruction capability, but provide limited generalizability ~\cite{Shin_2018_CVPR}. On the other hand, camera-frame reconstruction methods offer superior generalizability at the expense of reconstruction accuracy.  A novel method and representations have been proposed for category-level object pose estimation~\cite{Wang_2019_CVPR}. 

Inspired from a recent work~\cite{irshad2022centersnap} 
which learns the object pose and object geometry disjointly, we derive a method that preserves the reconstruction accuracy expected from object-frame reconstruction, but also provides a transformation to map the reconstruction to the camera frame using the predicted pose. This camera frame reconstruction is crucial for robot action planning.

\subsection{Grasp Estimation}
Grasp planning has also been one of the well-studied problems in robotics and continuous progress is still being made to make it more generalizable and robust. Early research on grasp planning focused on analytical methods for stable grasp predictions assuming known object geometries~\cite{FerrariCanny92, Roa14graspq}. In the last few years, learning-based approaches have shown impressive results for predicting grasps on previously unseen objects. Although most of the learning-based grasp estimation was focused on 2D grasp predictions in top-down view scenarios~\cite{mahler17dexnet2, Zeng22ARC}, recent methods also study the problem of grasp predictions in 6DoF~\cite{tenPas17gpd, mousavian20196, sundermeyer2021contact, breyer2021volumetric}. Many of these methods demonstrate impressive results, however, their performance is  often observed to be view-dependent. The grasp prediction accuracy of these methods is adversely affected when evaluated on cluttered scenes with occlusions and observed from a single view providing only partial geometric information of the objects in the scene.

\subsection{Simultaneous Shape Reconstruction and Grasp Estimation}
Scene understanding and grasp planning have been extensively studied in the last few decades and significant progress has been made in both the areas as discussed above. However, the interdependence on geometric scene reconstruction and robot action planning was mostly unexplored up until recently. In the last few years, researchers have studied the problem of simultaneous scene reconstruction and grasp planning~\cite{varley17shape, merwe19gag, jiang2021synergies, chavan2022simultaneous}. They demonstrate that the grasps planned with implicit understanding of underlying full 3D shape reconstruction are more accurate and avoid false positive grasps compared to other state-of-the art grasp planning methods which work on a single view of the scene. Zeng et al.~\cite{zeng2020transporter} and \cite{Zhanpeng2023} show that an implicit scene representation can guide placement-aware grasp planning for object rearrangement tasks. 

Inspired from these works, we propose a fast method which generates simultaneous shape and grasp predictions. On top of this, our method also provides explicit knowledge of object identity and pose which can be used for task-aware object manipulation.
\section{METHOD}
Given a single view RGBD image, our goal is to simultaneously perceive novel object instances, their full 3D shape, scale and pose, and their dense grasp parameters at real-time inference speed. Formally, given an RGBD image ($I \in \mathbb{R}^{h_0 \times w_0 \times 3}$, $D \in \mathbb{R}^{h_0 \times w_0}$) containing arbitrary $K$ novel object instances, we estimate a list of size $K$. Each element in the list contains (a) object category $c \in C$, where $C$ is the set of pre-defined categories, (b) pose $P \in SE(3) \equiv {\textbf{t} \in \mathbb{R}^3, r \in SO(3)}$, (c) scale $S \in \mathbb{R}$, (d) shape pointcloud $\textbf{Z} \in \mathbb{R}^{N \times 3}$ where $N$ is the predefined number of points, and (e) dense grasp parameters $G \in \mathbb{R}^{N \times 8}$ for every point in the shape pointcloud $\textbf{Z}$ as described in \secref{sec:scale_ae}.


We achieve these goals by combining the strengths of state-of-the-art categorical 3D perception with dense-grasp-parameter estimation. Specifically, we build on CenterSnap \cite{irshad2022centersnap}, a state-of-the-art categorical perception system where they input an RGBD image and predict scale, pose and shape of novel object instances.


An important feature of the CenterSnap network architecture that we maintain is the separation of the task of object geometry estimation and  the rest of the meta information (i.e., object category, pose, and scale). Following this decomposition of the problem, we first train a Scale-based Shape-Grasp Auto-Encoder (\SSGAE{}) to learn the low-dimensional latent combined space of shapes in unit-canonical space and scale dependent grasp distribution, i.e., every object is centered, of unit scale, and in canonical class orientation \cite{chang2015shapenet}. Then, we train a secondary network \SSGNET{} to regress object category, pose, object scale, and also the shape-grasp embedding. Using the shape-grasp embedding and the trained \SSGAE{}, we obtain full 3D pointcloud of the objects as well as dense per point grasp prediction in the unit-canonical space. The predicted object pose and scale further allows us to transform the predicted object pointcloud and grasps into the true-scale scene and in the camera frame.

Next, we describe \SSGAE{} in \secref{sec:scale_ae} and \SSGNET{} in \secref{sec:ssgnet}.



\subsection{Scale-based Shape-Grasp Auto-Encoder} 
\label{sec:scale_ae}

\begin{figure}
    \centering
    \includegraphics[width=1.0\columnwidth]{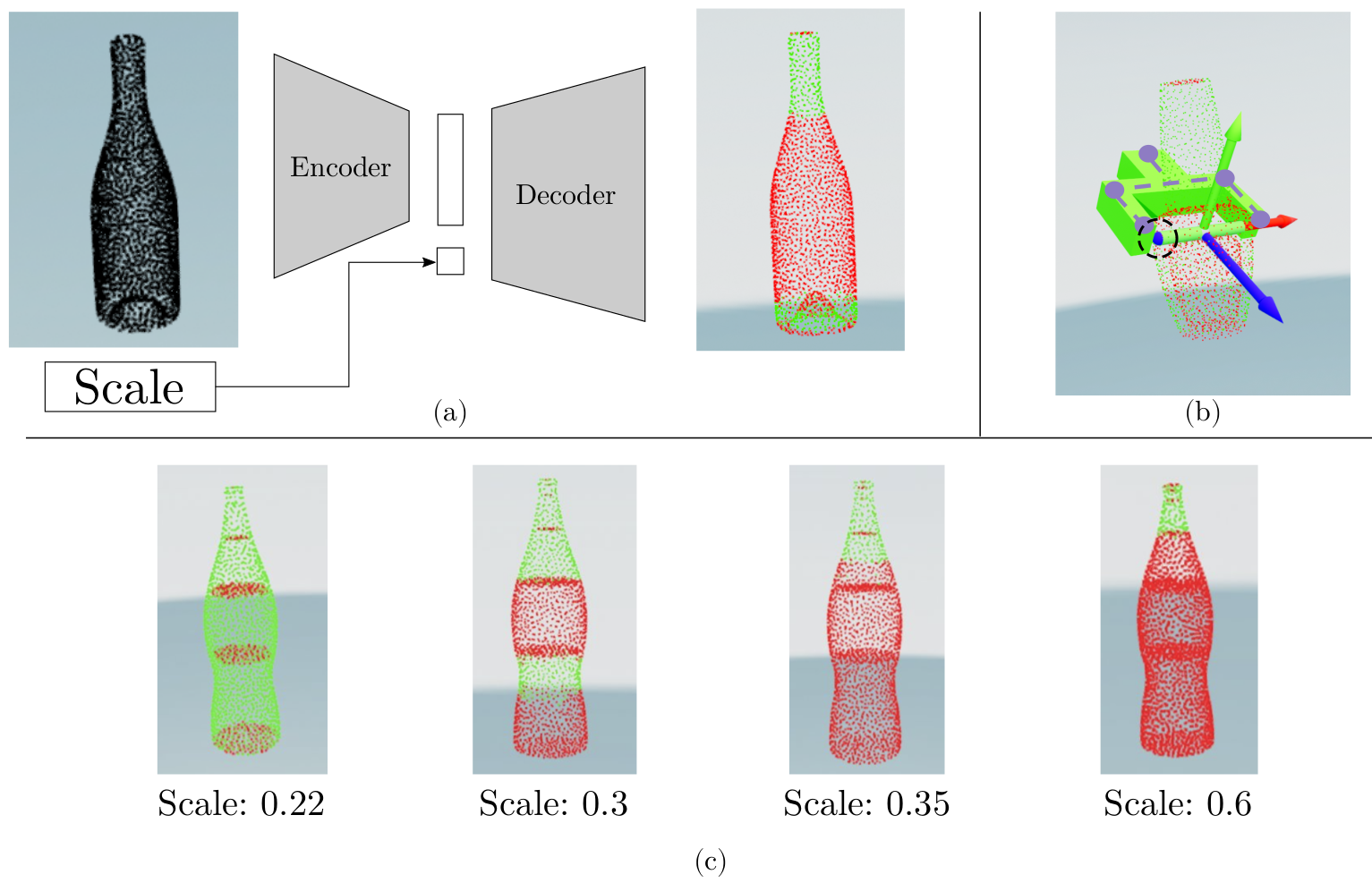}
    \caption{\SSGAE{} for learning combined latent space of shapes and scale dependent dense grasp parameters. (a) shows the encoder-decoder architecture (b) shows our per-point grasp-representation (c) shows the effect of scale on grasp-success predictions without changing the shape. Notice that as the scale increases, the wider parts of the bottle becomes un-graspable (red points) due to fixed maximum gripper width and only thinner parts remain graspable (green points).}
    \label{fig:scale_ae}
\end{figure}

The goal of scale-based shape-grasp auto-encoder is to learn a combined latent space of an object shape and scale dependent grasp parameters for the object. Our encoder decoder network takes the input of a pointcloud ($N \times 3$) in unit scale and canonical orientation, and scale ($s \in \mathbb{R}$) and learns to reconstruct the input pointcloud along with dense per-point grasp parameters described below.


\textbf{Grasp representation}: We follow the grasp representation proposed in~\cite{sundermeyer2021contact} (see \figref{fig:scale_ae} (b)), where every point $\textbf{p}$ in the pointcloud is labelled with grasp-success $gs \in [0,1]$, approach direction $\textbf{a} \in \mathbb{R}^3$, baseline direction $\textbf{b} \in \mathbb{R}^3$, and gripper width $gw \in [0, w_{max}]$ where $w_{max}$ is the maximum gripper opening. The final grasp pose $G \equiv \{\textbf{t}\in \mathbb{R}^3; R \in SO(3)\}$ can be obtained from these parameters as following ($d_0$ is the fixed distance between the gripper frame and the gripper base):
\begin{equation} \label{eq:grasp_pose}
    \textbf{t} = \textbf{p} + \frac{gw}{2} \times \textbf{b} - d_0 \times \textbf{a}
\end{equation}
    
\begin{equation}
\textbf{R} = \begin{bmatrix}
    \vert & \vert & \vert \\
    \textbf{b} & \textbf{a} \times \textbf{b} & \textbf{a} \\
    \vert & \vert & \vert
\end{bmatrix}    
\end{equation}


\textbf{Dataset}: We use the meshes from the NOCS categorical 6DoF perception dataset \cite{wang2019normalized} ($\approx 1200$ meshes) and choose $20$ different scales sampled from the scale distribution for each object category. The scale-distributions are obtained from the ShapeNet-Sem dataset \cite{savva2015semantically}. Each scaled mesh is then labelled with dense grasp parameters using our geometric grasp data generation tool. The final training dataset contains 24k datapoints.

\textbf{Architecture}: We use a PointNet-based encoder-decoder architecture \cite{qi2017pointnet}, where the encoder $E$ takes in the  $N \times 3$ pointcloud in the unit-canonical space and outputs a $128$-dimensional embedding. We append the object scale $s$ to this embedding and pass the $(128+1)$-dimensional vector to decoder $D$. Our decoder is a set of fully-connected layers which upsamples the embedding dimension to $N\times(3+17)$. This vector is then reshaped into a tensor of shape $N\times(3+17)$ where every point-vector is then processed by different heads. The first three elements are simply returned as point's 3D location. The fourth element is  applied  with sigmoid non-linearity to output the grasp-success confidence. The next three elements constitute $\textbf{z}_1$, and the three after that constitutes $\textbf{z}_2$. Approach direction and baseline direction are then obtained using $\textbf{z}_1$ and $\textbf{z}_2$ from Gram-Schmidt method as following:
\begin{equation}
   \textbf{ b} = \frac{\textbf{z}_1}{||\textbf{z}_1||}, \qquad  \textbf{a} = \frac{\textbf{z}_2 - <\textbf{b}, \textbf{z}_2> \textbf{b}}{||\textbf{z}_2||}
\end{equation}

The last 10 elements are then processed with soft-max layer and constitute one-hot grasp-width binned in equal segments across $[0, w_{max}]$, where $w_{max}$ corresponds to the maximum grasp-width. The final grasp-width is the width of the highest score bin. Note that the grasp-widths are predicted in the original gripper scale since they are estimated with the input scale in consideration, unlike the object pointcloud which is being predicted in the unit-canonical space as input.


\textbf{Training}: Training simultaneously for shape and scale dependent grasp parameter prediction is a challenging task due to their interdependence. Moreover, since there are no point-correspondences between predicted and ground truth pointcloud, estimating grasp-labels for loss calculation is non-trivial. We tackle this problem by our novel training paradigm where we back-propagate grasp-parameter losses only when the shape prediction loss reaches below a set threshold. To find grasp labels for predicted pointcloud, we interpolate the ground-truth grasp label from the point which are very close to the predicted point.

\textbf{Loss computation}: For a given data-point, we compute total loss, $L = \lambda_{\text{shape}} L_{\text{chamfer}} + \lambda_{\text{grasp}} L_{\text{grasp}}$, where $L_{\text{chamfer}}$ is the bidirectional Chamfer loss between input pointcloud $P^\ast$ and predicted pointcloud $\hat{P}$. If the Chamfer loss is greater than a threshold $\theta_{\text{chamfer}}$, then $L_{\text{grasp}} = 0$. Otherwise, $L_{\text{grasp}} = \lambda_{\text{gs}}L_{\text{gs}} + \lambda_{\text{gw}}L_{\text{gw}} + \lambda_{\text{6DoF}}L_{\text{6DoF}}$. Next, we describe what each of these loss terms refers to.

We define a list of nearest neighbor indices $nn_{\theta}(P_1, P_2) = [(i^{p}_{1}, j^{q}_{1}), (i^{p}_{2}, j^{q}_{2}) \cdots ]$ such that for every point $p$ in $P_1$, if a point $q$ exist in $P_2$ such that $||p - q||_2 < \theta$, then $(p,q)$ would be inside the list $nn_{\theta}(P_1, P_2)$. For brevity, we will just use $nn(P_1, P_2)$.

$L_{gs}$ captures loss for grasp-success prediction and is computed as:
\begin{equation*}
    \begin{split}
        L_{gs1} &= \text{top\_k}(\text{BCE}(\hat{P}[nn(\hat{P}, P^{\ast})[:, 0]].gs, P^{\ast}[nn(\hat{P}, P^{\ast})[:, 1]].gs)) \\
        L_{gs2} &= \text{top\_k}(\text{BCE}(P^{\ast}[nn(P^{\ast}, \hat{P})[:, 0]].gs, \hat{P}[nn(P^{\ast}, \hat{P})[:, 1]].gs)) \\
        L_{gs} &= mean(L_{gs1}, L_{gs2})
    \end{split}
\end{equation*}

where BCE stands for the Binary Cross-Entropy loss. We use the $p.gs$ notation to denote the grasp-success of a point $p$. Note that top\_k represents the top-$k$ elements with largest loss contributions. We use $k = 512$ in our experiments. This is done to handle class imbalance between success and failed grasp points.

\begin{figure}[t]
    \centering
    \includegraphics[width=\columnwidth]{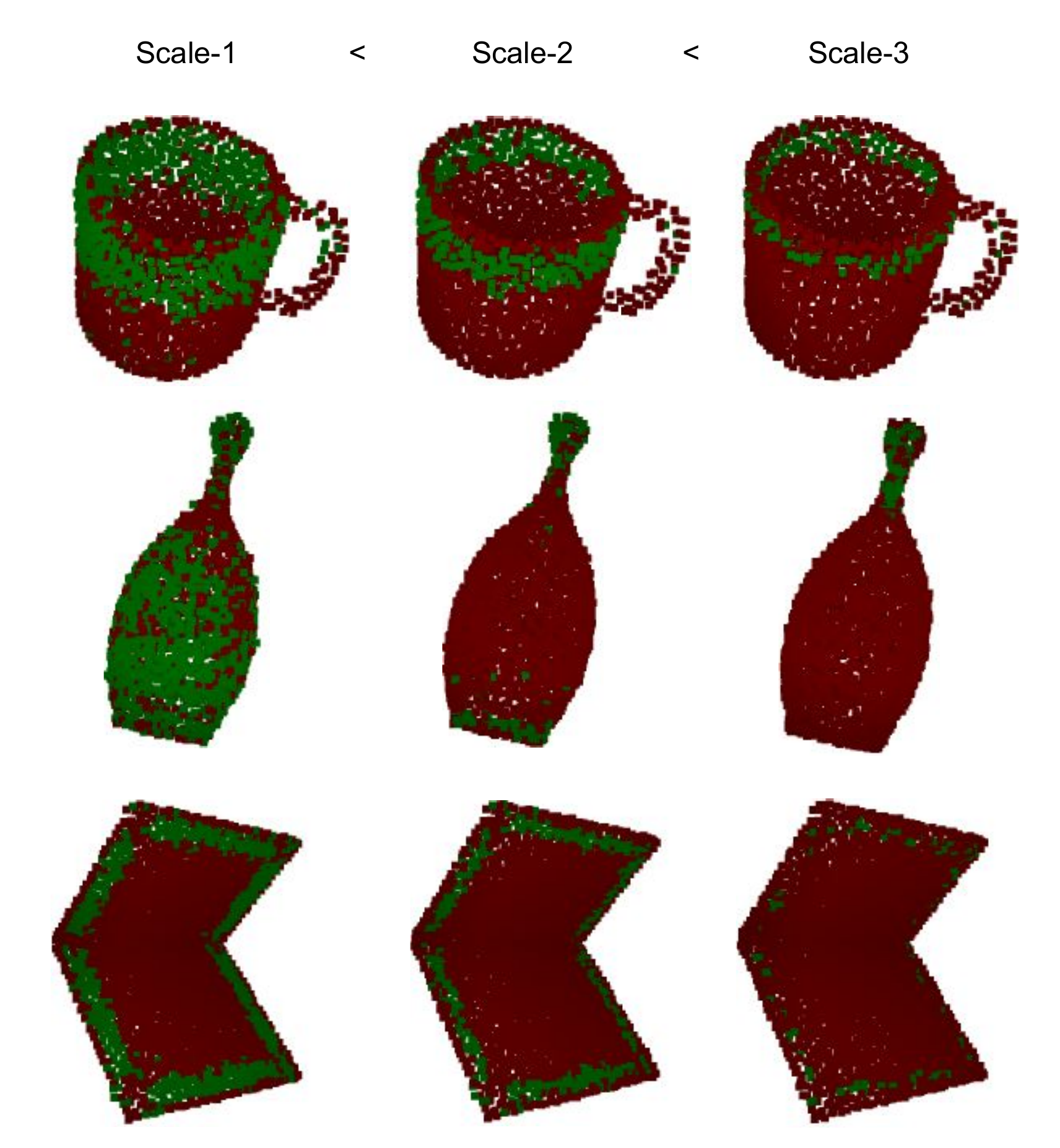}
    \caption{\textbf{\SSGAE{} Qualitative Results:} This figure demonstrates our network's ability to adjust grasp predictions based on the scale of the object. As the objects get larger in scale from left to right, and the predicted feasible grasps change accordingly. Feasible grasps points are depicted in green, and unfeasible in red.}
    \label{fig:scale_ae_results}
\end{figure}

The gripper-width and gripper-orientation losses are only computed only for estimated points that are close to ground-truth successful grasp points. Formally, we define the set of ground-truth successful grasp points as $P^{\ast+} \equiv \{p | p.gs=1,  \forall p \in P^{\ast}\}$.
Then the gripper-width loss $L_{gw}$ is computed as following: 
\begin{multline*}
    L_{gw} = \text{BCE}(\hat{P}[nn(\hat{P}, P^{\ast+})[:,0]].gwoh, \\
    P^{\ast+}[nn(\hat{P}, P^{\ast+})[:, 1]].gwoh)
\end{multline*}
\noindent where $p.gwoh$ refers to grasp-width one-hot prediction for a point $p$.  

For grasp-orientation prediction, we combine the predicted point $\hat{p}$, $a$, $b$, $gw$ to first estimate grasp pose (Eq. \ref{eq:grasp_pose}). Using grasp pose, we estimate the positions for gripper keypoint locations (Fig. \ref{fig:scale_ae}b). The loss $L_{\text{6DoF}}$ is distance loss between these predicted keypoints and ground truth keypoints. Since the gripper is symmetric, we also compute loss with symmetric keypoints.

\begin{figure*}[t]
    \centering
    \includegraphics[width=\linewidth]{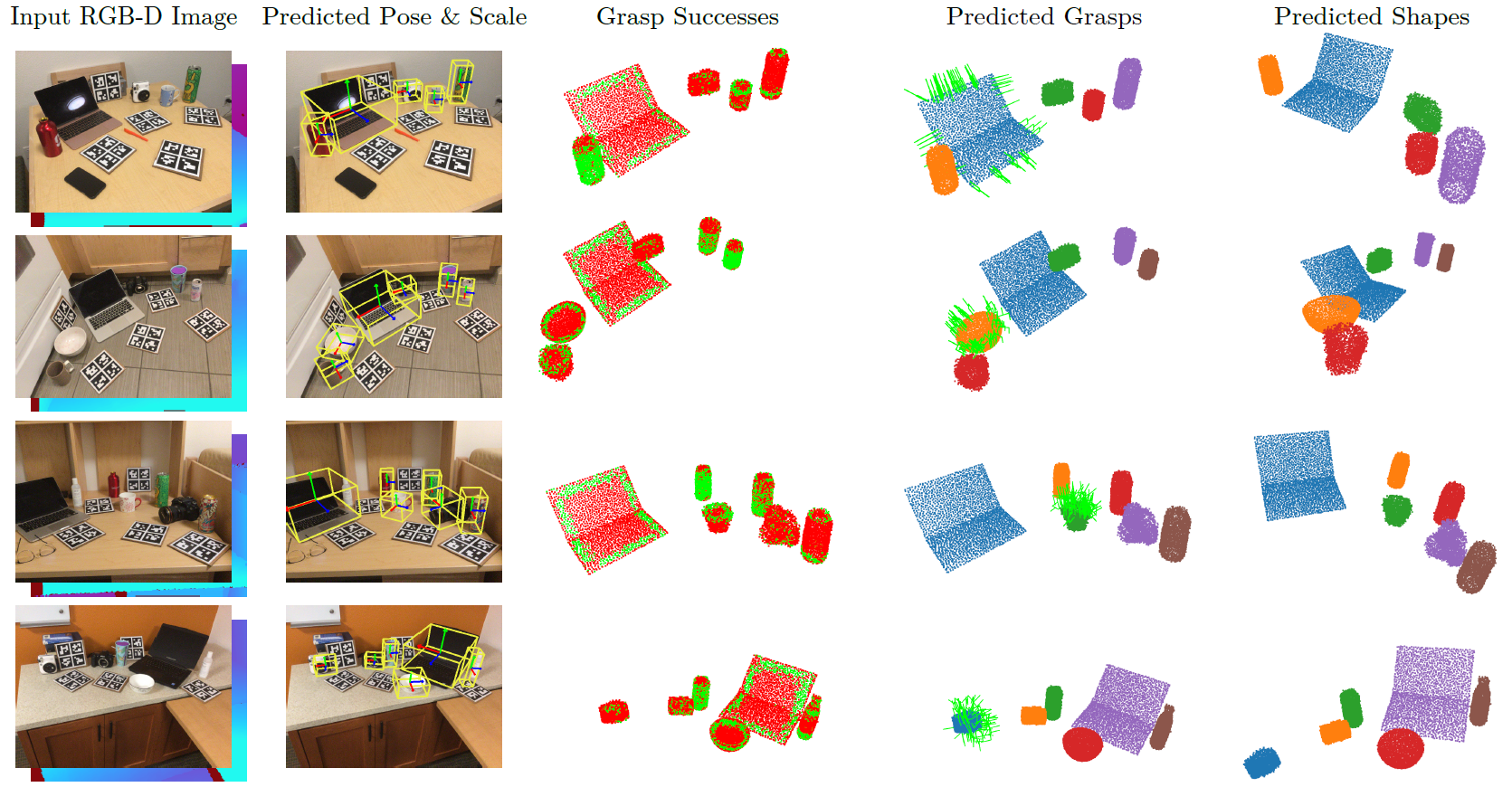}
    \caption{\textbf{Qualitative results.} These qualitative results demonstrate our method's ability to predict quality shapes and grasps. From columns left to right: The input RGBD image, the predicted pose and scale of each object shown as 3D bounding boxes, the predicted success of grasping each predicted point with successful in green and unsuccessful in red, the predicted grasps (grasps only visualized for one object per scene for clarity, but predicted for all), and finally the predicted shapes of all the objects.}
    \vspace{-3mm}
    \label{fig:qual-results}
\end{figure*}

\subsection{\SSGNET{}}\label{sec:ssgnet}
Given an RGBD image, the goal of \SSGNET{} is to predict 6DoF pose, scale, and embedding for \SSGAE{}'s decoder (\figref{fig:method}). The architecture and the training of \SSGNET{} closely follow centersnap model, where given an RGBD image, we predict a heatmap representing object centers and corresponding dense object descriptors for all pixels in the heatmap. Dense object descriptors are sampled at local peaks in the heatmap which represents our object detection. Each object descriptor is passed seperately through the rest of the network to obtain object pose, scale, shape and grasp parameters.

\textbf{Object center heat-map detection}
To localize the objects in the input image, we implement the object center heat-map prediction. In order to optimize for speed while maintaining high accuracy, we utilize a ResNet\cite{DBLP:journals/corr/HeZRS15} based network to take in the RGBD image and predict a heat-map with peaks at object centers. During inference, peak detection is performed over this heat-map to localize object centers in the image. This heat-map is defined as $M \in \mathbb{R}^{\frac{h_0}{8} \times \frac{w_0}{8} \times (1 + z)}$, where the $\frac{h_0}{8}$ and $\frac{w_0}{8}$ represent a reduced resolution from the input image. For each pixel in the heat-map, the network predicts a single confidence value correlating to an objects center, as well as a dense object descriptor $z$ representing the networks predictions for the pose, scale, and shape/grasp embeddings.

\textbf{Dense object descriptor}
Dense object descriptor $z$ is a 1D vector containing estimations for object pose, scale, and shape/grasp embeddings. Object pose and scale values are used as it is from the embedding after appropriate vector reshape. The scale value is appended to the shape/grasp embeddings before being passed through pre-trained \SSGAE{}, resulting in the predicted shape and dense grasp predictions.

During training, the network is supervised using the ground truth object pose, scale and \SSGAE{}'s embedding which is obtained after passing ground-truth shape to \SSGAE{}'s encoder.

\section{Experiments}
In this section, we evaluate our method for the following capabilities (a) Perception - 3D object-detection, 6D pose estimation, and shape reconstruction (b) Dense grasp estimation, and (c) end-to-end inference time.

\textbf{Dataset}:
We used the NOCS dataset for all our experiments. NOCS dataset is a collection of RGBD images, containing objects from six categories: bottle, bowl, can, mug, laptop and camera. The dataset also provides 6DoF poses for all object instances present in the image. We further label each object instance in the dataset with dense grasp labels using the tool in~\cite{chavan2022simultaneous}. We follow the provided train-test splits where we train our models using CAMERA-train set, fine-tune them on REAL-train set and evaluate on REAL-test set. The shape-grasp auto encoder is trained only on models from CAMERA-train dataset. \figref{fig:qual-results} shows the qualitative results of our method on a few representative examples from REAL-test set. Our method is able to predict the object shapes and grasps accurately. Moreover, the estimated object poses provide a way to transform the reconstructed objects to the camera frame and compose the complete scene. We discuss the detailed quantitative evaluation of our method below.

\subsection{Perception}
3D object detection and 6D pose estimation - we independently evaluate 3D object detection and 6D pose estimation using following key metrics: (a) Average-precision for various IOU-thresholds, (b) Average precision for object instances for which the error is less than $n^{\circ}$ rotation and $m$ cm translation. 3D Shape reconstruction is evaluated using the Chamfer distance.

We compare our method with the following baselines: (1) Normalized Object Coordinate Space (NOCS) - uses a Mask-RCNN like architecture to predict NOCS maps, and combine NOCS predictions with observed depth to predict pose and size; (2) DeformNet -  predicts per-object 2D bounding box, and deforms a mean shape to get final shape reconstruction; (3) CenterSnap - predicts objects as peaks in heatmap, where every peak is augmented with dense embedding for absolute pose, scale and shape embedding which is further passed to shape decoder to obtain shape.

\begin{figure}
    \centering
    \includegraphics[width=\columnwidth]{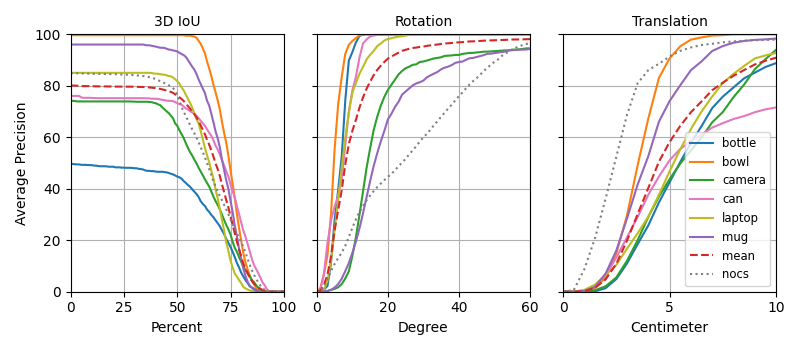}
    \caption{mAP vs 3D IoU, Rotation error and Translation error plots on NOCS-Real test set. From the plots we observe that the pose and size estimates from out method are comparable to the state-of-the art method NOCS\cite{wang2019normalized}. The object orientation prediction performance of our method is better than NOCS; however, the object position prediction is comparatively less accurate. Still, we get the position accuracy of about 3 cm for 80\% of the test cases.}
    \label{fig:mAP}
\end{figure}

%
3D object detection and 6D pose estimation results are reported in \tabref{tab:pose} and \figref{fig:mAP}. We note that our method is able to outperform the NOCS baseline in all metrics except IOU25, where NOCS even performs better than the current SOTA CenterSnap. 

3D shape reconstruction results are reported in \tabref{tab:chamfer} where we report the bidirectional Chamfer distance metric. Our method shows comparable performance to CenterSnap. On average, our method has $1.8$ cm Euclidean distance error which is under the tolerance limit of robotics task of our interests.

\begin{figure}[t]
    \centering
    \includegraphics[width=\columnwidth]{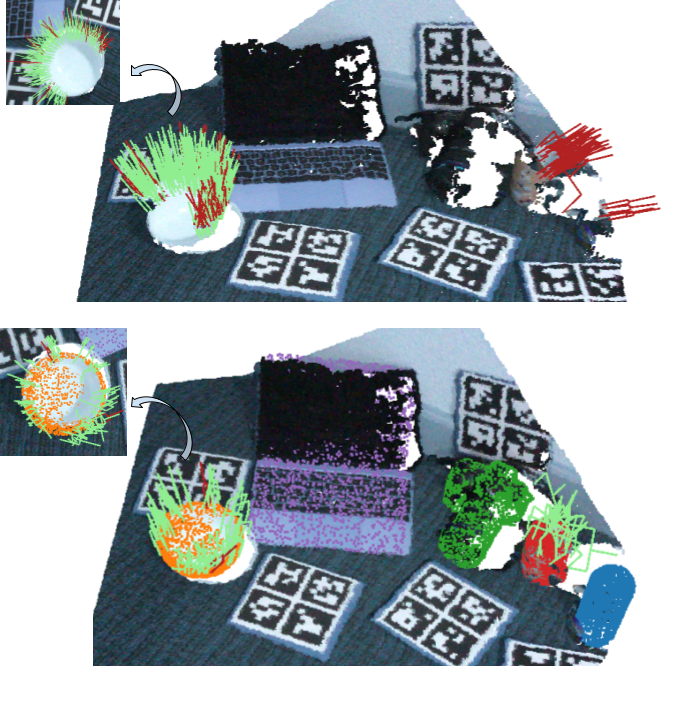}
    \caption{Comparison of grasp prediction accuracy of \OURS{} (bottom) with ContactGraspNet (top) \cite{sundermeyer2021contact}. Thanks to the joint shape-grasp training, our method is able to provide much more accurate grasps on the object while avoiding false positive grasp proposals.}
    \label{fig:cgn-us}
\end{figure}

\begin{table}[t]
\centering
 \begin{tabular}{l | c c c c c c} 
 \hline
 \textbf{Method} & IOU25 & IOU50 & 5°5cm & 5°10cm & 10°5cm & 10°10cm \\
 \hline
 NOCS & \textbf{84.8} & 78.0 & 10.0 & 9.8 & 25.2 & 25.8 \\
 CenterSnap & 83.5 & \textbf{80.2} & \textbf{27.2} & \textbf{29.2} & \textbf{58.8} & \textbf{64.4} \\
 Ours & 79.7 & 76.2 & 18.4 & 22.5 & 37.5 & 54.9 \\
 \hline
 \end{tabular}
 \caption{NOCS 3D pose estimation results: Evaluated using IOU precision with different thresholds and average precision for pose of instances with thresholds in degrees for rotation and cm for translation.}
 \label{tab:pose}
\end{table}

\begin{table}[t]
\centering
 \begin{tabular}{l | c  c  c } 
 \hline
 \textbf{Object} & DeformNet\cite{tian2020shape} & CenterSnap\cite{irshad2022centersnap} & Ours \\
 \hline
 Bottle & 0.50 & \textbf{0.13} & 0.19 \\
 Bowl & 0.12 & 0.10 & \textbf{0.09} \\
 Camera & 0.99 & 0.43 & \textbf{0.41} \\
 Can & 0.24 & \textbf{0.09}  & \textbf{0.09} \\
 Laptop & 0.71 & \textbf{0.07}  & 0.13 \\
 Mug & 0.10 & \textbf{0.06} & 0.17 \\
 \hline
 \end{tabular}
 \caption{NOCS 3D shape reconstruction results: Evaluated with Chamfer distance metric ($10^{-2}$ m). Lower Chamfer score indicates better reconstruction accuracy.}
 \label{tab:chamfer}
\end{table}

\figref{fig:scale_ae_results} shows qualitative evaluation of \SSGAE{} where the model is not only able to reconstruct fine geometry details, but also able to capture the variation of grasp parameters as the scale changes.

\subsection{Dense Grasp Estimation}

\begin{table}[h!]
\centering
 \begin{tabular}{l | c c c c c} 
 \hline
 Coverage rate & 10\% & 20\% & 30\% & 40\% & 50\% \\
 \hline
 Success rate & 92.3 & 89.7 & 87.3 & 84.2 & 78.3 \\
 \hline
 \end{tabular}
 \caption{Grasp success rate for different coverage rates. The trend observed in the table shows that as more (but less confident) grasps are sampled to achieve larger coverage, grasp success reduces as expected.}
 \label{tab:succ-cov}
\end{table}

A typical metric for the accuracy of grasp prediction is grasp success rate. The grasp success rate is defined as a ratio of the number of successful grasp trials to the number of attempted grasp trials. 
However, the success rate can change under different environmental constraints such as occlusions in a cluttered environment or kinematic reachability the robots. A dense grasp prediction provides more feasible grasp candidates to choose the best grasping pose under these environmental constraints. 


To evaluate the diversity of predicted dense grasps, we measure the coverage-rate. A ground truth grasp is considered covered if a predicted gripper-base of the predicted grasp is within 2cm distance of the gripper-base of the ground truth grasp. Coverage rate is number of ground truth grasps covered divided by the total number of ground truth grasps. Following~\cite{sundermeyer2021contact}, we report success-rate for different coverage-rates in \tabref{tab:succ-cov}. We observe that \OURS{} is able to maintain consistently high grasp success rate while being able to cover high percentage of ground truth grasps. \figref{fig:cgn-us} shows qualitative comparison with Contact-Grasp-Net \cite{sundermeyer2021contact}.

 
\subsection{Inference time}
The major advantage of our method is fast inference for dense grasp prediction. In this section, we compare \OURS{} end-to-end inference time with the  reported inference time of various grasping baselines: (1) Contact-Graspnet~\cite{sundermeyer2021contact}: 4 FPS (2) “Volumetric
Grasping Network (VGN) \cite{breyer2021volumetric}: 22 FPS, (3) Grasp detection
via Implicit Geometry and Affordance (GIGA) \cite{jiang2021synergies}: 22 FPS.
The inference time of \OURS{} is \textbf{\FPS{} FPS} on a standard computer with NVIDIA GTX 1080 Ti (11.2GB) graphics card.



The inference time of \OURS{} is significantly lower than the inference times of baselines, while providing several advantages compared to the baselines. VGN inputs a fused pointcloud from multiple camera view and this muti-view fused point-cloud creation is not included in their inference time computation. Collecting multiple images from different view points using a camera on a robot is a very slow process and can lead to view alignment errors.
GIGA outputs predict grasps as well as full scene reconstruction from a single-view RGBD input. However, they do not detect object instances, and their 6DoF poses. On the other hand, \OURS{} is able to significantly improve the inference time as compared to these baselines, and is able to detect object instances, their 3D shape, and 6DoF poses from a single RGBD image. 

\section{CONCLUSION}

In this paper we presented \OURS{}, a real-time method for simultaneous object-level scene understanding and grasp prediction. Specifically, given an RGBD image of a cluttered scene, \OURS{} generates a class label, a complete 3D reconstruction, scale, pose and  dense grasp predictions for each object.

The two-step structure and training of \OURS{} is key to its performance: First, we train an object-scale-dependent auto-encoder to output object reconstruction and dense grasp proposals in a unit-sized canonical object frame. Training in this canonical frame provides our method superior reconstruction capability which directly contributes to accurate grasp prediction. Second, we train a network to generate semantic segmentation, object sizes, object poses, and object embeddings for the shape-grasp decoder. This semantic information allows transforming the decoded object shape and grasps to the camera frame for robotic applications such as grasping and motion planning.

With extensive comparisons with baselines, we demonstrate that \OURS{} is a on par with the baselines while providing real-time inference (\FPS{} FPS) and semantic scene information needed for downstream tasks. Such reactivity and detailed scene understanding is essential for robots to perform task-driven manipulation in dynamic environments. In our future work, we would like to generate task-level information such as grasps informed by the downstream task~\cite{Zhanpeng2023}.







\bibliographystyle{IEEEtran}
\bibliography{IEEEabrv,refs}
\end{document}